# An Explainable, Attention-Enhanced, Bidirectional Long Short-Term Memory Neural Network for Joint 48-Hour Forecasting of Temperature, Irradiance, and Relative Humidity

Georgios Vamvouras[1], Konstantinos Braimakis[1], Christos Tzivanidis[1]

[1]Laboratory of Refrigeration, Air-Conditioning and Solar Energy,

School of Mechanical Engineering, National Technical University of Athens, Greece

## Abstract

This paper presents a Deep Learning (DL) framework for 48-hour forecasting of temperature, solar irradiance, and relative humidity to support Model Predictive Control (MPC) in smart HVAC systems. The approach employs a stacked Bidirectional Long Short-Term Memory (BiLSTM) network with attention, capturing temporal and cross-feature dependencies by jointly predicting all three variables. Historical meteorological data (2019–2022) with encoded cyclical time features were used for training, while 2023 data evaluated generalization. The model achieved Mean Absolute Errors of 1.3 °C (temperature), 31 W/m² (irradiance), and 6.7% (humidity), outperforming state-of-the-art numerical weather prediction and machine learning benchmarks. Integrated Gradients quantified feature contributions, and attention weights revealed temporal patterns, enhancing interpretability. By combining multivariate forecasting, attention-based DL, and explainability, this work advances data-driven weather prediction. The demonstrated accuracy and transparency highlight the framework's potential for energy-efficient building control through reliable short-term meteorological forecasting.

| Nomenclature | | | |
|---|---|---|---|
| AWS | Amazon Web Services | NWP | Numerical Weather Prediction |
| BiLSTM | Bidirectional Long Short-Term Memory | relhum | Relative Humidity |
| CMA-ES | Covariance Matrix Adaptation Evolution Strategy | SARIMA | Seasonal Auto-Regressive Integrated Moving Average |
| DL | Deep Learning | temp | Temperature |
| GRNET | Greek Research and Technology Network | WRF | Weather Research and Forecasting model |
| HVAC | Heating, Ventilation, and Air Conditioning | GFS | Global Forecast System |
| IG | Integrated Gradients | irrad | Solar Irradiance |
| LSTM | Long Short-Term Memory | month_sin | Sine-encoded month (cyclical feature #1) |
| MAE | Mean Absolute Error | month_cos | Cosine-encoded month (cyclical feature #2) |
| ML | Machine Learning | solh_sin | Sine-encoded solar hour (cyclical feature #3) |
| MPC | Model Predictive Control | solh_cos | Cosine-encoded solar hour (cyclical feature #4) |

## 1. Introduction

Accurate short-term weather forecasting is necessary for a wide range of applications, such as renewable energy management and efficient Heating, Ventilation, and Air Conditioning (HVAC) system control. The main meteorological variables that affect energy consumption in buildings and occupant thermal comfort are temperature, relative humidity and solar irradiance [1, 2]. Precise and efficient Model Predictive Control (MPC) requires accurate forecasting of those variables over short horizons, spanning up to 48 hours into the future [3].

Numerical Weather Prediction (NWP) models, such as the Global Forecast System (GFS) [4] and the Weather Research and Forecasting (WRF) model [5], are commonly used for medium-range forecasting. Studies report 48-hour ahead temperature MAEs around 1.5 °C for GFS and WRF [6]. For solar irradiance, WRF-Solar achieved close to 23% relative MAE in a 48-hour forecast focusing on Austria, whereas in other cases, WRF reached relative MAEs of 25% and GFS around 28% [7]. Despite their accuracy, NWP models often are computationally expensive and do not offer localized forecasts, necessary for building control.

Machine Learning (ML) methods have also been explored recently, as potential alternatives. Long Short-Term Memory (LSTM) networks have been successfully applied to weather time-series forecasting, as they are capable at capturing complex temporal dependency patterns. In a notable study over the Greek island of Crete [8], LSTM, SARIMA, and hybrid SARIMA-LSTM models were applied for local forecasts, achieving MAEs of around 2.1 °C for temperature and 10% for relative humidity at 48 hours, where hybridization showed marginal improvements. Different studies across other climates reached MAEs of around 5.8%, limited however to a 24-hour horizon [9]. It should be noted that most existing DL approaches focus on a single meteorological variable at a time, with only few exceptions [10], thus not accounting for cross-variable correlation which plays a critical role in precision.

Furthermore, many prior studies overlook the aspect of model explainability, and even though attention mechanisms have been previously employed in other research fields, aiming to enhance the learning of temporal dependency, they have rarely been applied to meteorological BiLSTM models. In a similar manner, techniques such as Integrated Gradients (IG) are quite rare, even though they offer significant potential in inferring feature importance in weather forecasting [11].

To address these gaps, this study introduces a BiLSTM architecture with an attention mechanism, for joint 48-hour ahead forecasting of temperature, solar irradiance, and relative humidity. Unlike existing models, the proposed approach jointly predicts all variables, capturing their physical interdependencies. An attention mechanism dynamically weights inputs of historical weather information based on timestep-level significance, while Integrated Gradients provide feature-focused explanations of predictions. The model is evaluated on a multi-year dataset (2019–2023) provided by the NASA POWER API [12] and demonstrates 48-hour MAEs of approximately 1.3 °C for temperature, 31 W/m² for irradiance, and 6.7% for relative humidity. These results are comparable to, and even surpass, current state-of-the-art NWP and DL approaches. The combination of joint multivariate forecasting, attention-based learning, and explainability establishes this model as a promising tool for predictive HVAC control and for advancing meteorological forecasting.

## 2. Methodology

### 2.1. Data Selection

In HVAC control, the meteorological variables that have the most profound effect on building energy consumption and load, are the external temperature of the surrounding air, relative humidity and solar irradiance. Specifically, as shown in recent studies, temperature is the single most significant factor which affects building cooling and heating loads [13], as it explains the majority of variance in energy consumption, typically above 70% [14]. As outdoor temperature declines, heating demand rises in order to maintain indoor comfort, whereas when it rises, cooling demand increases due to significant air-conditioning use. Apart from temperature, relative humidity is often considered the second most influential meteorological variable, especially in hot and humid climate conditions [8]. Increased humidity leads to higher latent cooling load, as HVAC systems need to maintain indoor comfort by removing it from the air. Finally, solar irradiance is also a major contributor to cooling loads due to heat gains from building envelopes [15], most notably so during the summer months.

For all these reasons, the latter meteorological variables were selected for joint prediction in this study, accompanied by cyclical time-related features, aiming to capture solar time and month-based dependencies in variable predictions. To retain the periodicity of solar time and months, each of the two features was encoded using sine and cosine waves, according to the following formula:

$$x_{sin} = \sin\left(\frac{2\pi}{T}t\right), x_{cos} = \cos\left(\frac{2\pi}{T}t\right) \quad (1)$$

where $t$ is the current timestep (month index or solar hour) and $T$ the period (12 for months and 24 for solar hour). Without cyclical encoding, for instance, months like January and December are close in temporal context, contrary to their serial numeric representations, 1 and 12 respectively, which are far apart. Another critical advantage of this transformation is that the final feature values belong in the interval [-1, 1], which is already properly scaled for use as model input.

Overall, the final input is a time series which consists of $N_p$ past and $N_f$ future timesteps of 7 features each. In each past timestep, the feature vector comprises temperature, irradiance, relative humidity, month sine, month cosine, solar hour sine and solar hour cosine. However, in each future timestep, the first three variables are unknown, and are replaced by zeros, while the latter four are kept and used to enhance temporal context.

**2.2. Model Architecture**

As meteorological data exhibits strong temporal patterns, models that can capture seasonal and diurnal cycles as well as long-range dependencies, such as LSTMs, are well suited for accurate forecasting. However, standard LSTMs process input only in the forward direction, from past to future, which limits their ability to fully exploit the future cyclical input features, namely month and solar hour sine and cosine. In contrast, Bidirectional LSTM layers process sequences in both forward and backward directions [16, 17], enabling the model to integrate contextual information from the entire input window and more effectively capture temporal patterns that span across the forecasting horizon.

Traditional BiLSTM layers, on the other hand, do not distinguish between the importance of each timestep, even though recent timesteps have greater influence on the final predictions. Adding an attention mechanism mitigates this issue, as it allows the model to dynamically weigh the contribution of each timestep, focusing on the most important parts of the sequence. This improves forecast accuracy, especially for longer horizons, by preventing irrelevant historical data from diluting the learning signal [18]. During training, a dropout layer was added after each BiLSTM layer in order to prevent overfitting during training, by deactivating a fixed percentage of randomly selected neurons.

Additionally, a custom attention mechanism was added, operating upon the output of the final BiLSTM layer, shaped as [batch size, timesteps $(= N_p + N_f)$, 2×BiLSTM units (=16)]. An attention score is computed for each individual timestep by passing it through two fully connected (Dense) layers, of size 64 and 1 respectively, which is afterwards normalized using Softmax. Then, the BiLSTM outputs are multiplied by these weights. Attention provides interpretability by exposing the learned weighting patterns, offering insight into which timesteps the model relies on the most when generating predictions. This is particularly valuable for smart HVAC applications, where decision-makers benefit from transparent forecasting models.

It should be noted that during the output stage, after passing through the attention mechanism, the model utilizes a time-distributed fully connected layer. It applies the same transformation using the identical weights to each timestep individually, thereby preserving the temporal structure of the sequence without collapsing it into a single vector, as would a traditional fully connected layer.

For the final selection of the model's hyperparameters as well as training-specific parameters, Covariance Matrix Adaptation Evolution Strategy (CMA-ES) was used, as implemented in Python's cma package [19]. The latter is a population-based, evolutionary optimization algorithm designed for continuous, non-linear, and non-convex optimization problems, particularly effective in model fine-tuning and hyper-parameter optimization [20]. The objective function of the optimization process was the sum of per-timestep MAE of each feature, scaled by its range for comparability. The algorithm was tasked to jointly optimize the values of $N_p$, learning rate, dropout rate, number of BiLSTM layers and their size. It concluded to the following combination, seen in Table 1. Model architecture is depicted in Figure 1.

*Table 1: Final model hyperparameters after CMA-ES optimization*

| Hyper-parameter | Optimized value |
|---|---|
| **Number of past timesteps ($N_p$)** | 22 |
| **Learning rate** | 0.0031 |
| **Dropout rate** | 0.053 |
| **Number of BiLSTM layers** | 2 |
| **Size of each BiLSTM layer** | 8 forward + 8 backward = 16 total |

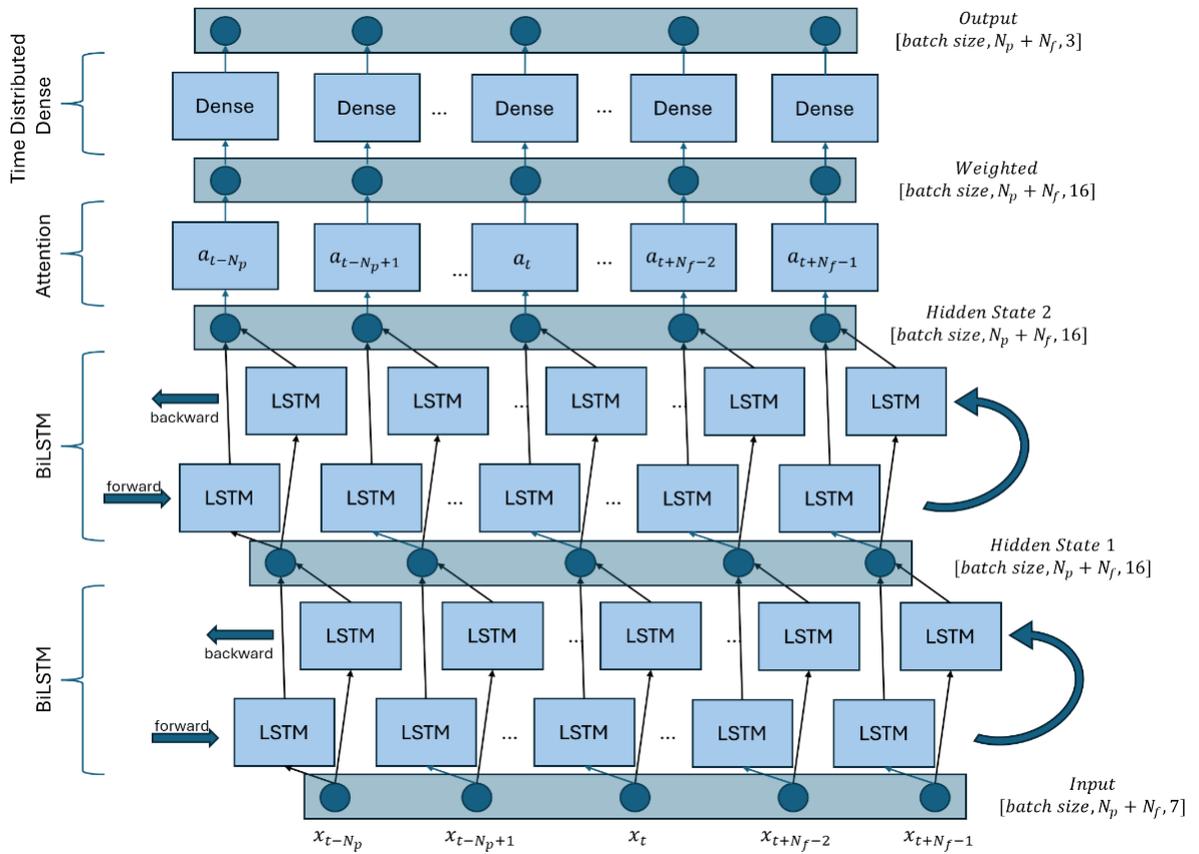

*Figure 1: Model architecture.*

A batch size of 64 was selected for acceleration, as training and testing sample sequences were 34993 and 8693 respectively. Min-Max scaling was performed individually only for each of the meteorological features (temperature, irradiance and relative humidity), as the four cyclical features inherently take values in the interval [-1, 1], which is an appropriate range for model input. Naturally, the saved scalers were applied to the testing input sequences, to avoid data leakage.

This work utilized computational resources provided by Amazon Web Services (AWS) through the Greek Research and Technology Network (GRNET) in the context of the strategic partnership with the National Technical University of Athens (NTUA). Specifically, a g5.24xlarge AWS virtual machine was used, which consists of 96 vCPUs, 384 GiB of RAM, 4 NVIDIA A10G Tensor Core GPUs (each with 24 GB of GPU memory) and 200 GB of storage.

## 3. Results and Discussion

After training the model on sequences corresponding to the years 2019 up to and including 2022, performance was evaluated on the 2023 test dataset. The model achieved a MAE of 1.32 °C for temperature, 31.51 W/m² for irradiance, and 6.71% for relative humidity on the 2023 test set, as shown in Table 2. These values correspond to 7%, 15%, and 10% of each variable's standard deviation, respectively. Note that the MAE for relative humidity is reported in absolute percentage points (%), not as a relative error. In addition, the per-timestep MAE is shown in **Figure 2**. As expected, the MAE for all features is lower at approximately the first seven timesteps, because they are closer to the present and more directly influenced by recent historical input features. At later timesteps though, the MAE rises with a slow but steady rate, with the exception of solar irradiance, where the rise is slightly more pronounced. This can be attributed to the naturally unpredictable nature of irradiance, as it strongly depends on transient cloud cover, which is itself a chaotic and highly variable phenomenon, even on an hourly basis. On the contrary, temperature and relative humidity vary more gradually, due to thermal inertia and higher dependency on daily cycles. Figure 3 shows the actual predicted time series for a sample of the test dataset, 1, 25 and 48 hours before occurrence, in blue, orange and green respectively, alongside the ground truth values shown in black. It can be observed that even though expectedly there are differences between the predictions and the true values, there is excellent alignment of the local extrema. This suggests that the cyclical features used in the input sequences, allowed the model to infer the meteorological features' periodicity with astonishing precision, which is crucial in applications that require knowledge of the time of day when meteorological variables are expected to be at a minimum or maximum, such as smart HVAC control and energy trading.

Figure 4 illustrates the average attention weights assigned to each input timestep by the model when forecasting future values. The horizontal axis denotes the position of the hourly timesteps relative to the present, with negative values referring to past timesteps and positive values referring to future timesteps. The vertical axis shows the mean attention weight across all test samples, with the shaded region indicating ±1 standard deviation, and the vertical dashed line marks the boundary between past and future input windows. Notably, the model allocates the highest attention to the final timestep of the historical input, nearest to the present, while further onto the past, values drop significantly. However, for future timesteps the weights are mostly stable around 0.0175, with the exception of the last 8 timesteps which influence the model more noticeably more. This suggests that the model depends less on past timesteps the further they are from the present, while it focuses more on recent ones. In addition, future timesteps exert similar influence, as they comprise uniform cyclical temporal features without the inherent variability associated with meteorological features. The peak at the end of the sequence could either be an artifact, or an indication of the diminishing influence of past meteorological features balanced by increased weight on the cyclical features.

Aiming to enhance model explainability, Integrated Gradients [21] were used to interpret the model's decision process by attributing forecast outcomes to specific input features, collectively and for specific timesteps. The contribution of each input feature to the forecasts, averaged over all timesteps and test sequences, is illustrated in Figure 5. The most important feature is the relative humidity, followed closely by temperature and then by solar irradiance. This observation is consistent with expectations, as relative humidity tends to change gradually, depending on sustained evaporation or condensation as well as factors such as ground moisture, vegetation, and large-scale air mass movement. Moreover, relative humidity affects temperature through latent heat transfer. Evidently, these properties render it a dependable predictor of its own future values as well as of future temperature. In a similar manner, temperature also exhibits more gradual changes compared to other variables, due to the high thermal inertia of the atmosphere and Earth's surface, which buffer rapid energy fluctuations and smooth out temperature variations over time.

In contrast, solar irradiance changes rapidly with the movement of clouds, the position of the sun, or weather events, because it is a direct measure of the energy arriving at the surface and is not stored in the same way as the previous two meteorological variables. Despite that, solar irradiance affects temperature, as it is the primary energy input to the surface and atmosphere, gradually warming them throughout the day. Overall, even though solar irradiance is crucial for temperature prediction, its rapid fluctuations and inherent unpredictability render it a less reliable predictor relative to temperature or relative humidity. The diagram also reveals that the cyclical features provide essential information about the underlying periodicity and diurnal cycles in the data. Their contribution is particularly important for future prediction horizons, where recent meteorological values become less informative. They enable the model to account for regular, recurring patterns, thereby improving forecast accuracy and robustness.

Figure 6 demonstrates the contribution of each input feature to the prediction of temperature and how it varies among timesteps. According to the graph, relative humidity and temperature itself are the most influential predictors, with attribution scores at around 0.009, which then decrease and remain mostly stable at 0.007, before beginning to decline further at the last 10 timesteps. The model has successfully inferred the significance of recent input to the immediate future predictions, thus adapting to rely on them more heavily. On the other hand, the temporal cyclical features have the least influence on temperature as they remain stable around 0.001. Irradiance occupies an intermediate position, with an attribution score close to 0.004, asserting moderate influence relative to the other two meteorological features.

Similarly, Figure 7 shows that the most significant predictor of solar irradiance is temperature, whose attribution score remains high until around the 15$^{th}$ timestep, rapidly declining thereafter. After the first two timesteps, irradiance and relative humidity, despite starting at values close to temperature, also exhibit a steep decline, eventually reaching attribution scores similar to those of the month and solar hour encodings. This indicates that the model relies heavily on temperature for the first few timesteps, and from then on cyclical temporal features dominate, which is consistent with the widely known unpredictability of irradiance due to its dependence on chaotic cloud movement and weather phenomena.

Finally, as far as relative humidity is concerned, Figure 8 shows that temperature and relative humidity itself are the most significant predictors, with relative humidity consistently outscoring the rest of the features. The attribution of relative humidity exhibits a steadily decreasing trend, reminiscent of exponential decay, with the highest importance at short prediction horizons and a gradual, slower decline over longer horizons. This behavior indicates that the model has learned to reflect the inertia of relative humidity, meaning that changes in atmospheric moisture content occur gradually over time. Significant shifts in humidity require sustained processes such as evaporation, condensation, or the movement of large air masses. As a result, recent relative humidity values are highly informative for short-term forecasts, but their influence naturally diminishes as these slow physical processes lead to only gradual changes over longer periods.

| Table 2: Testing MAE and test dataset statistics | | | | | | |
|---|---|---|---|---|---|---|
| **Feature** | **MAE** | **Units** | **Min** | **Max** | **Mean** | **Std** |
| **Temperature** | 1.32 | °C | 2.21 | 42.32 | 18.78 | 7.45 |
| **Solar Irradiance** | 31.51 | W/m² | 0.00 | 989.30 | 200.97 | 279.65 |
| **Relative Humidity** | 6.71 | % points | 16.82 | 100.00 | 68.60 | 18.84 |

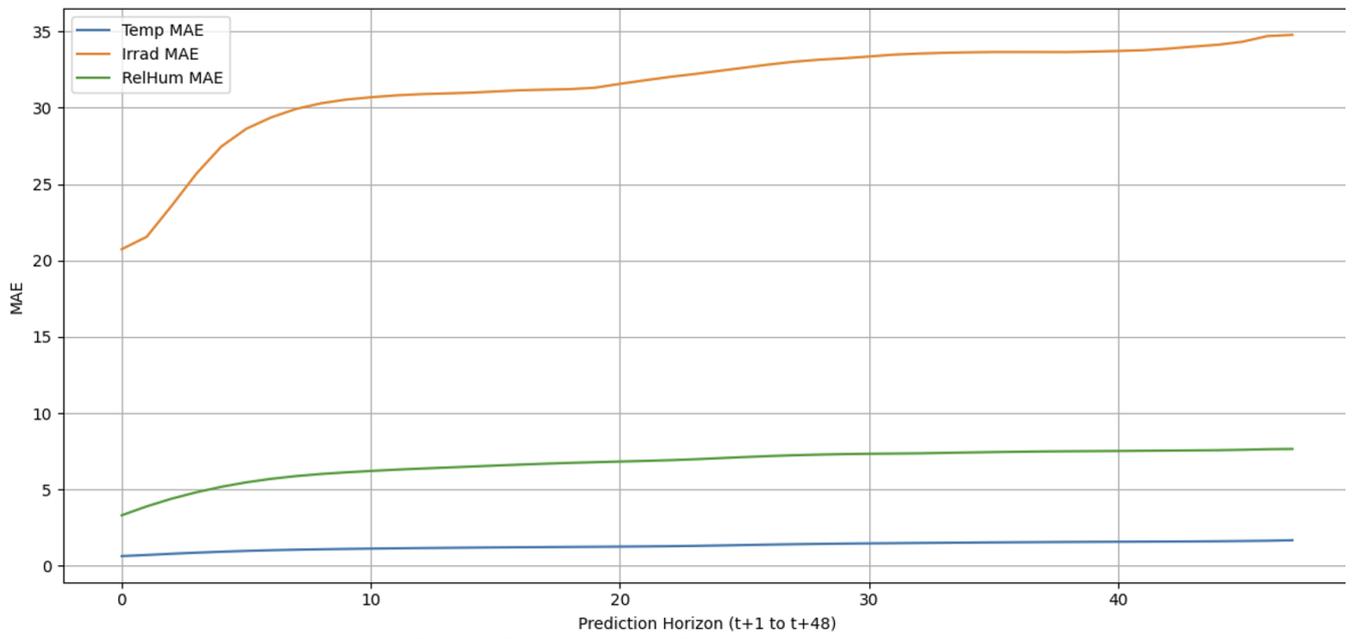

Figure 2: Per timestep MAE for each output feature

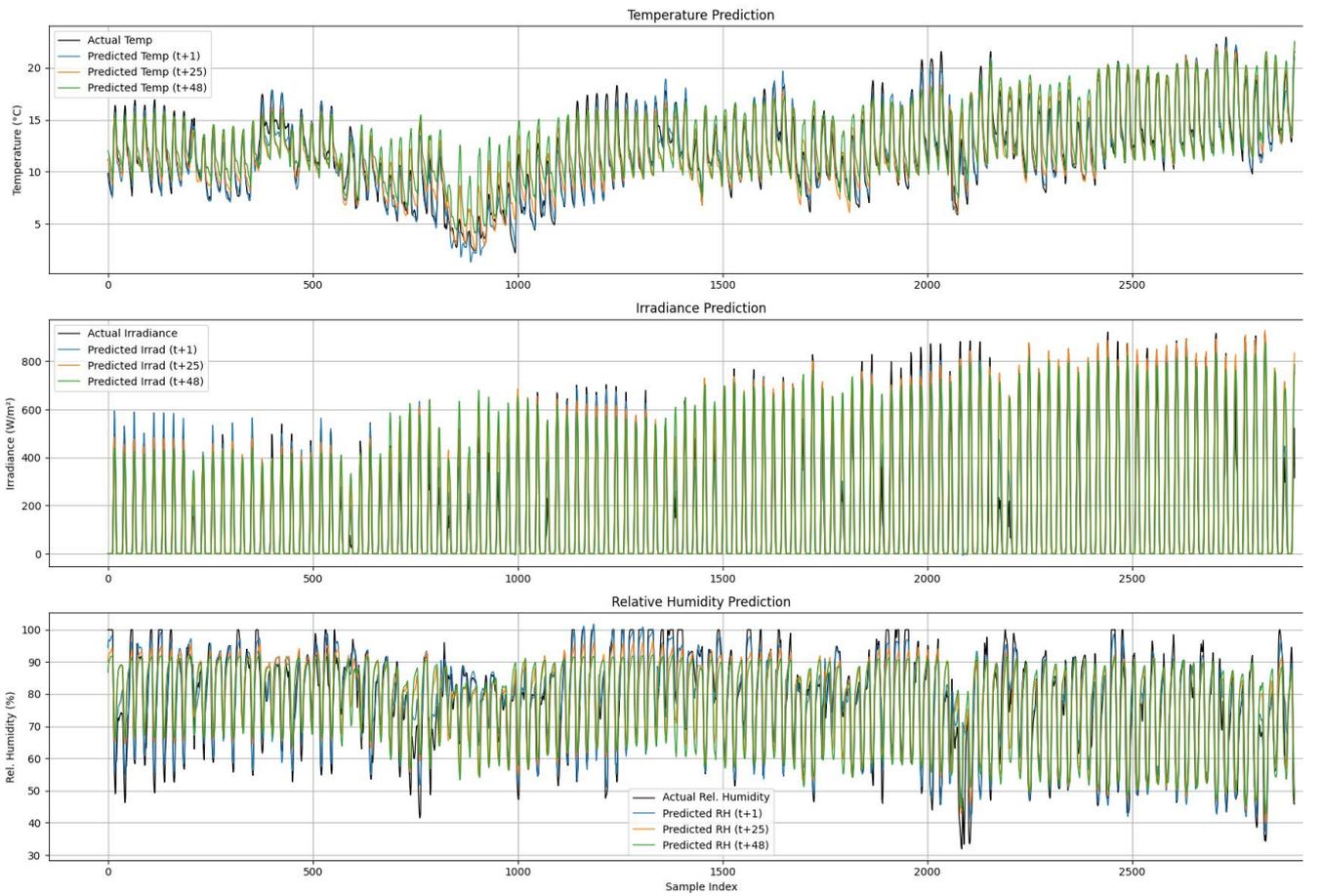

Figure 3: Comparison of predicted features made 1, 25 and 48 hours before occurrence, with the actual values.

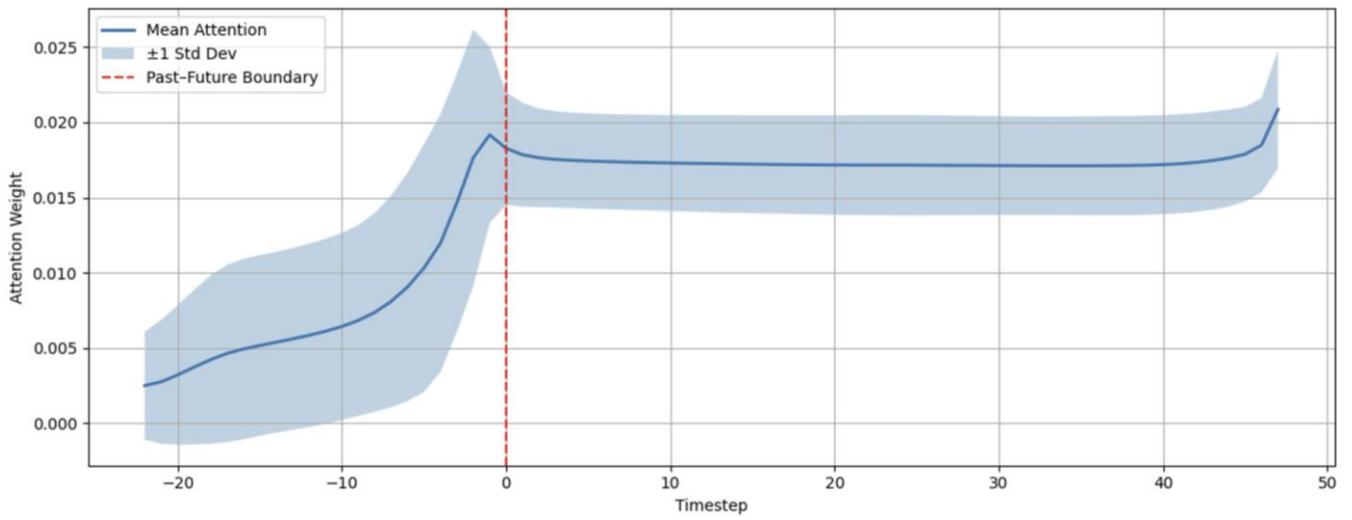

*Figure 4: Average attention weights assigned to each input timestep.*

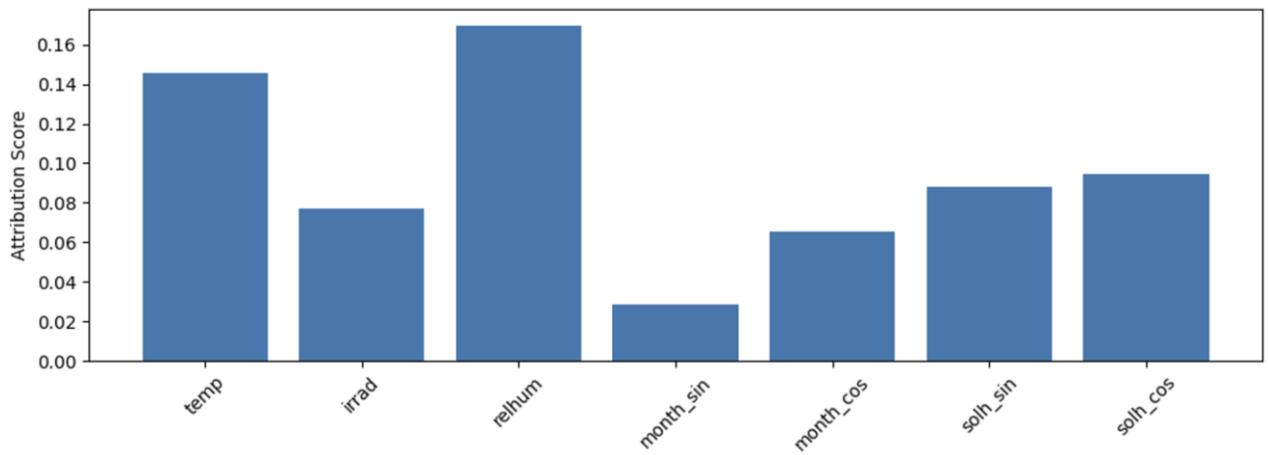

*Figure 5: Overall feature importance.*

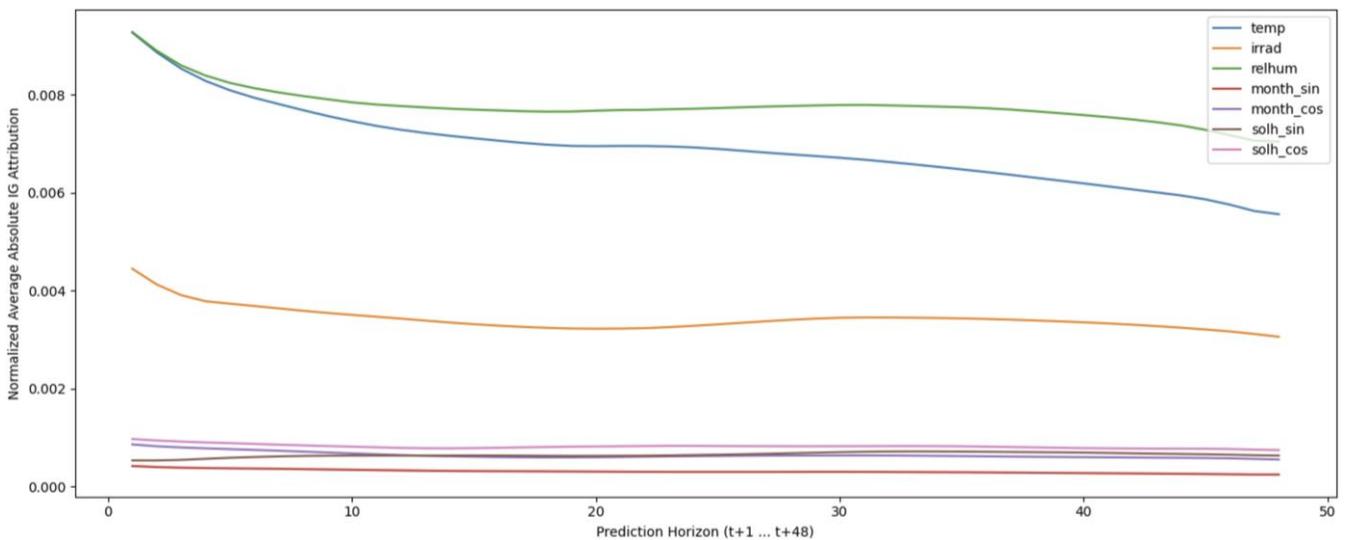

*Figure 6: Contribution of each input feature to the prediction of temperature.*

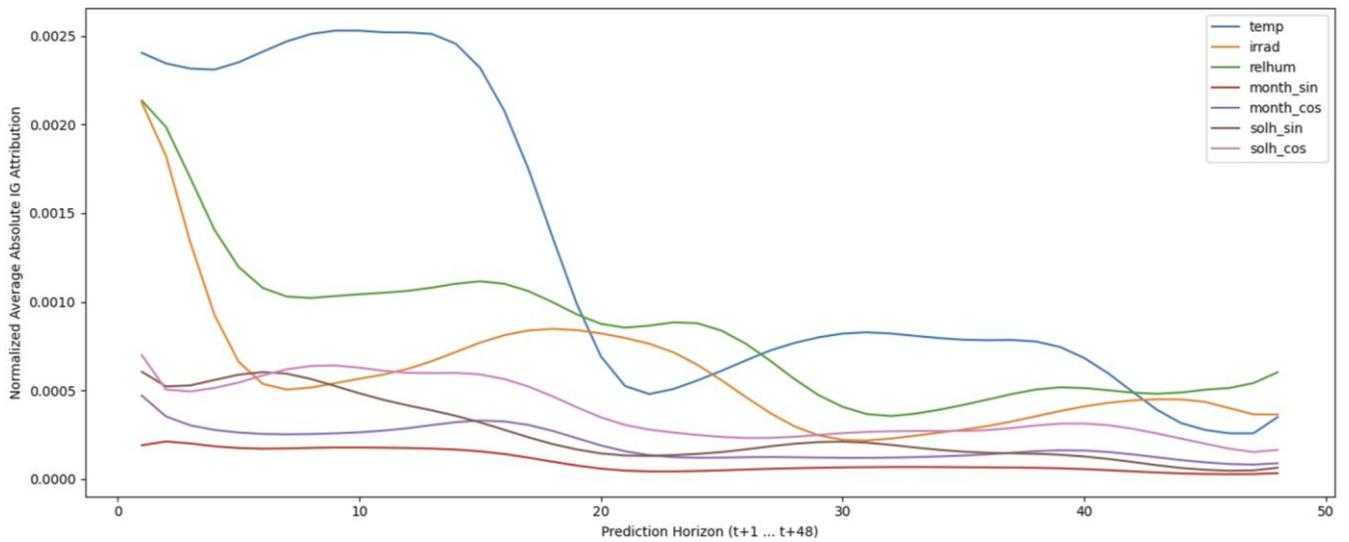

*Figure 7: Contribution of each input feature to the prediction of irradiance.*

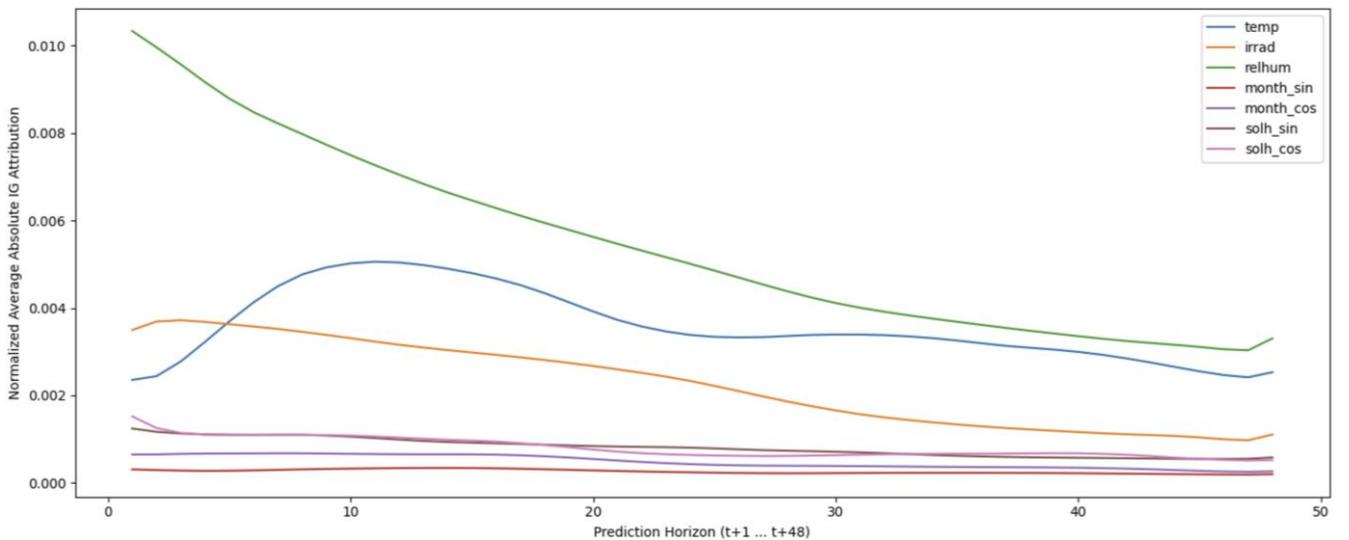

*Figure 8: Contribution of each input feature to the prediction of relative humidity.*

## 4. Conclusions

This work introduces an attention-enhanced BiLSTM model, trained on data from 2019 up to 2022 and tested on 2023 data, which successfully manages to jointly predict temperature, solar irradiance and relative humidity, for a forecast horizon of 48 hours. The model achieved MAE values of 1.32 °C for temperature, 31.51 W/m² for irradiance, and 6.71% for relative humidity, results that surpass current state-of-the-art statistical and ML/DL models reported in the literature, to the best of the authors' knowledge. Additionally, it takes advantage of the interdependence between the meteorological variables by simultaneously predicting all three of them, while the application of Integrated Gradients enhances model explainability and transparency, providing intuitive insights into how the model arrives at its predictions. These traits are particularly beneficial for applications like smart HVAC control, where accurate and explainable forecasts of multiple weather variables can support real-time system optimization, as well as for energy trading, where improved predictability and interpretability can enhance decision-making and risk management.